\documentclass[letterpaper]{article} 
\pdfoutput=1
\usepackage{aaai25}  
\usepackage{times}  
\usepackage{helvet}  
\usepackage{courier}  
\usepackage[hyphens]{url}  
\usepackage{graphicx} 
\usepackage{amsmath}  
\usepackage{amsfonts}
\usepackage{dsfont}
\usepackage{booktabs}
\usepackage{subcaption}
\usepackage{diagbox}
\usepackage{amssymb}  
\usepackage{natbib}  
\usepackage{caption} 
\usepackage{algorithm}
\usepackage{algorithmic}

\usepackage{newfloat}
\usepackage{listings}
\DeclareCaptionStyle{ruled}{labelfont=normalfont,labelsep=colon,strut=off} 
\floatstyle{ruled}
\newfloat{listing}{tb}{lst}{}
\floatname{listing}{Listing}
\title{Enhancing Decision Transformer with Diffusion-Based Trajectory Branch Generation}
\author{
    Zhihong Liu,
    Long Qian,
    Zeyang Liu,
    Lipeng Wan,
    Xingyu Chen,
    Xuguang Lan\thanks{Corresponding author}
}
\affiliations{
    Xi'an Jiaotong University, Xi'an, China, 710049\\

    liuzhihong@stu.xjtu.edu.cn, xglan@mail.xjtu.edu.cn
}

\usepackage{bibentry}
\begin{document}

\maketitle

\begin{abstract}
Decision Transformer (DT) can learn effective policy from offline datasets by converting the offline reinforcement learning (RL) into a supervised sequence modeling task, where the trajectory elements are generated auto-regressively conditioned on the return-to-go (RTG).
However, the sequence modeling learning approach tends to learn policies that converge on the sub-optimal trajectories within the dataset, for lack of bridging data to move to better trajectories, even if the condition is set to the highest RTG.
To address this issue, we introduce Diffusion-Based Trajectory Branch Generation (BG), which expands the trajectories of the dataset with branches generated by a diffusion model.
The trajectory branch is generated based on the segment of the trajectory within the dataset, and leads to trajectories with higher returns.
We concatenate the generated branch with the trajectory segment as an expansion of the trajectory.
After expanding, DT has more opportunities to learn policies to move to better trajectories, preventing it from converging to the sub-optimal trajectories.    
 Empirically, after processing with BG, DT outperforms state-of-the-art sequence modeling methods on D4RL benchmark, demonstrating the effectiveness of adding branches to the dataset without further modifications. 
\end{abstract}

%

\section{Introduction}
Offline Reinforcement learning (RL) ~\cite{intro_offlinebegin1,intro_offlinebegin2}, which learns effective policies entirely from previously collected data without directly interacting with the environment, has gained much attention. 
It has particularly wide application in scenarios where interacting with the environment using untrained policies can be costly or dangerous~\cite{intro_realworldcost1,intro_realworldcost2}.
The Decision Transformer (DT)~\cite{intro_DT} uses a transformer architecture to maximize the likelihood of actions conditioned on history trajectories and the RTG.
This approach transforms offline RL into a supervised sequence modeling task.
\begin{figure}[t]
    \centering
    \includegraphics[width=0.8\linewidth]{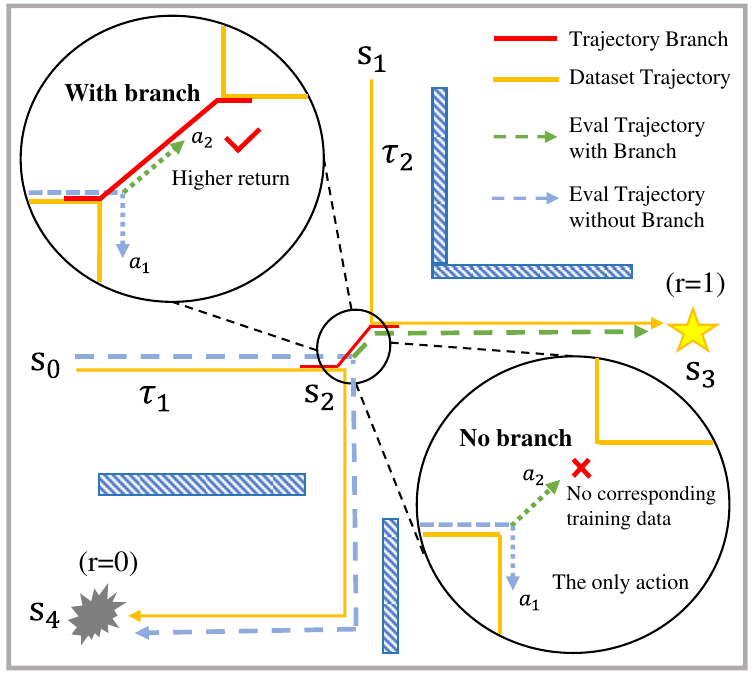}
    \caption{A maze example to illustrate the problem of DT converging to sub-optimal trajectories and the importance of trajectory branches. Eval Trajectory refers to the trajectory generated by the policy during evaluation.}
    \label{fig:intro}
\end{figure}

However, the sequence modeling approach lacks the ability to learn better policies that need to move across the trajectories. 
As previous work~\cite{intro_slproblem} has indicated, offline RL methods based on dynamic-programming can stitch together pieces of experience to solve tasks in a way that has not been explicitly experienced, but methods based on supervised-learning (SL) do not have an explicit mechanism for stitching.
SL-based RL methods fail to perform stitching even when trained on abundant quantities of data ~\cite{intro_slproblem}. 
Specifically for sequence modeling methods, this problem manifests as the sequence modeling approach tends to converge on the sub-optimal trajectories in the datasets, lack of the ability to learn better policies that need to move across the trajectories.
As shown in Fig.\ref{fig:intro}, in the case of no branch, agent can only learn to follow the sub-optimal trajectory, for it is the only sequence of transitions the agent has learned in this situation. 

DT as a representative sequence modeling method, a lot of works have been proposed to improve it, but they primarily focus on online fine-tuning or pre-training ~\cite{intro_ODT,intro_PretrainDT}. Some works are about stitching ability, but they focus on the RTG or the history length maintained by DT ~\cite{QDT,intro_EDT}.
The problem caused by DT's sequence modeling learning approach has received little attention.

To mitigate this problem, we expand the trajectories of the dataset with trajectory branches, which can prevent DT from converging on the sub-optimal trajectories. 
The branch leads to trajectories with higher returns, providing DT with more opportunities to learn policies that can move to better trajectories rather than follow the sub-optimal trajectory.
As shown in Fig.\ref{fig:intro}, in the case of with branch, agent can branch off the sub-optimal trajectory by the action learned from the branch, resulting in a better policy. 

To generate trajectory branches, we propose Diffusion-Based Trajectory Branch Generation (BG), which uses a diffusion model to generate trajectory branches based on segments of trajectories within the dataset.
We use the Trajectory Value Function (TVF) to guide the generation, to make the generated branches leads to trajectories with higher return.
Then we concatenate the branch with the trajectory segment as an expansion of the trajectory.
The generated branch provides DT with more opportunities to branch off the sub-optimal trajectory and learn better policies.

We evaluate BG on Gym, Maze2d, and Antmaze tasks from the D4RL benchmark~\cite{D4RL}. 
Our method significantly improves the performance of DT merely by expanding trajectories within the dataset with branches, particularly on tasks where transitioning from sub-optimal trajectories to optimal ones is more difficult.
After processing with BG, DT's performance becomes competitive with offline RL methods and outperforms state-of-the-art sequence modeling methods without further modifications.

\begin{figure*}[t]
\centering
\includegraphics[width=1\textwidth]{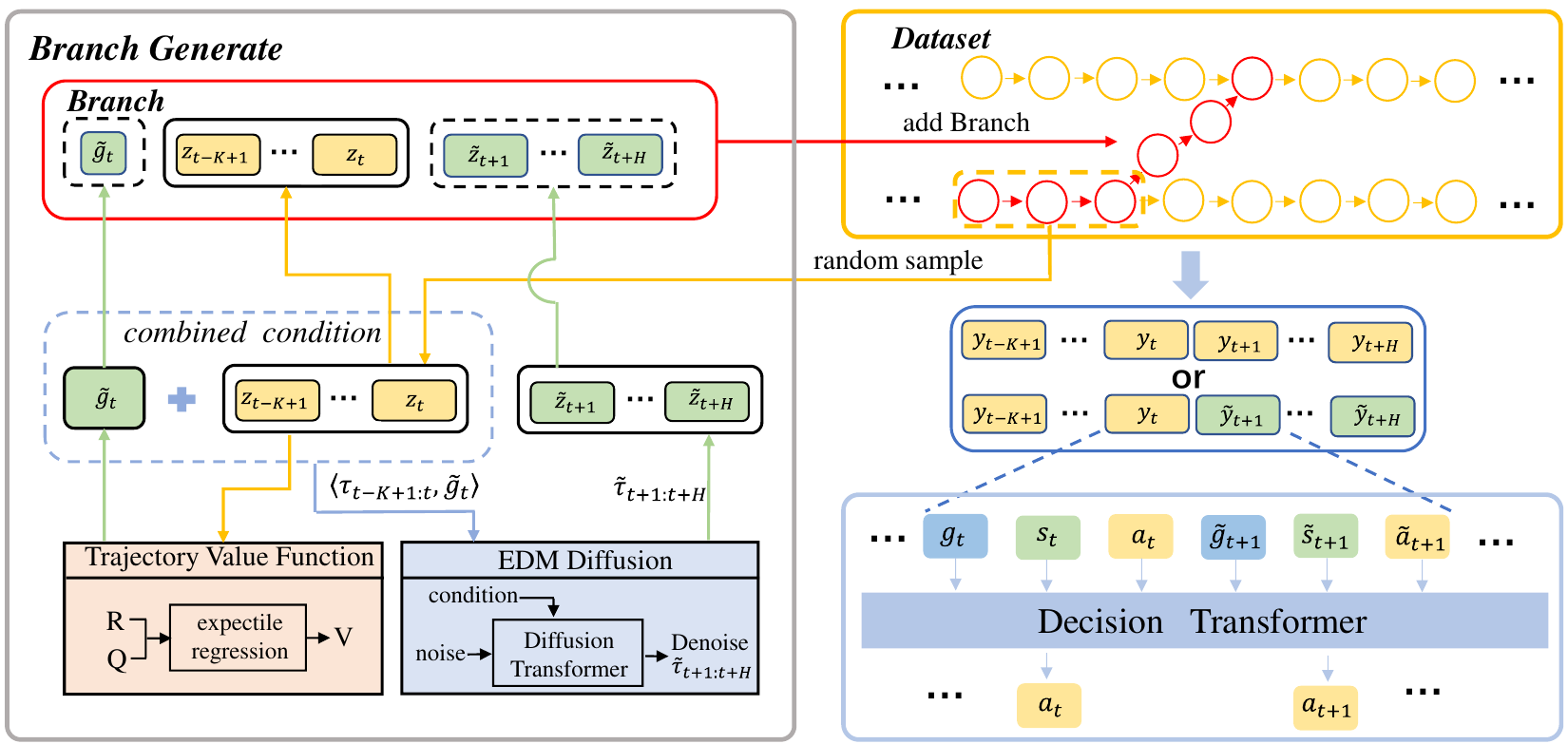} 
\caption{\textbf{Overall pipeline of BG}. The trajectory segments are randomly sampled from the trajectories of the dataset. The TVF generates $\tilde{g}_{t}$ based on the trajectory segment. The $\tilde{g}_{t}$ and the trajectory segment are combined together as the condition, then fed into the EDM diffusion model. The generated branches is concatenated to the trajectory segments as expansions of the trajectories in the dataset. Then DT is trained on the expanded dataset.}
\label{fig: method}
\end{figure*}

\section{Background}
\paragraph{Sequence Modeling in Offline RL} 
Offline RL learns policies on static dataset~\cite{relwork_offlinebegin1}. DT~\cite{intro_DT} formulates offline RL as supervised sequence modeling. The offline dataset can be denoted as a collection of trajectories 
\begin{math} D=\{\cdot\cdot\cdot,s_{t}^{(n)},a_{t}^{(n)},r_{t}^{(n)},\cdot\cdot\cdot\} \end{math},
 DT predicts the actions based on the previous trajectories concatenated with RTG $g_{t}^{(n)}$:
\begin{equation}
    \mathcal{L}_{DT} = \mathbb{E}_{t,n}\left[a_{t}^{(n)}-\pi_{DT}\left(\langle g,s,a \rangle_{t-K}^{(n)};g_{t}^{(n)},s_{t}^{(n)} \right)\right]^{2}
    \label{eq:train dt}
\end{equation}
where $E_{t,n}$ is an omission of $E_{t \in [0,T],n \in [1,N]}$. $g_{t}^{(n)}$ is the RTG defined as 
\begin{math} g_{t}^{(n)}\dot{=}\sum_{t'=t}^{T}r\left(s_{t'}^{(n)},a_{t'}^{(n)}\right) \end{math} and $\langle g,s,a \rangle_{t-K}^{(n)}$ denotes the previous $K$ timesteps trajectory concatenated with RTG $g_{t}^{(n)}$. The policy $\pi_{DT}$ is implemented by a transformer. For each timestep $t$, three different tokens $g_{t}^{(n)},s_{t}^{(n)},a_{t}^{(n)}$ are fed into the model, and the future action is predicted via auto-regressive modeling.

\paragraph{In-sample Learning via Expectile Regression} 
To avoid the out-of-distribution(OOD) actions,
IQL~\cite{IQL} uses only in-sample actions to learn the optimal Q-function. IQL uses an asymmetric $L_{2}$ loss (i.e.,expectile regression~\cite{expectile_regression1,expectile_regression2}) to learn the V-function, which can be seen as an estimation of the maximum Q-value over actions that are in dataset support:
\begin{equation}
    \begin{aligned}
        \mathcal{L}_{V} &= \mathbb{E}_{(s,a) \sim \mathcal{D}} [L_{2}^{\tau}(Q(s,a)-V(s))] \\
        \mathcal{L}_{Q} &= \mathbb{E}_{(s,a,s') \sim \mathcal{D}} [(r(s,a)+\gamma V(s') - Q(s,a))^{2}] 
    \end{aligned}
    \label{eq:iql_loss}
\end{equation}
where $L_{2}^{\tau}(u) = |\tau - \mathds{1}(u<0)|u^{2}$, and $\mathds{1}$ is the indicator function, $\mathcal{D}$ represents the dataset. After learning Q and V, IQL extracts the policy by advantage-weighted regression:
\begin{equation}
    \mathcal{L}_{\pi} = \mathbb{E}_{(s,a) \sim \mathcal{D}} [\exp(\beta (Q(s,a) - V(s)))\log{\pi(a|s)}]
\end{equation}

\paragraph{Score-Based Diffusion Models}
Diffusion models~\cite{diffusion1} are a class of generative models that generate samples by reversing a noising process.
We consider a diffusion process 
$\{\boldsymbol{x}^{\tau}\}_{\tau \in [0,\mathcal{T}]}$, where $\tau$ is a continuous time variable.
The corresponding marginals are 
$\{p^{\tau}\}_{\tau \in [0,\mathcal{T}]}$,
with boundary conditions 
$p^{0}=p^{data}$ and $p^{\mathcal{T}}=p^{prior}$,
where $p^{prior}$ is a tractable unstructured prior distribution.
The denoise process can be described as the solution to a standard stochastic differential equation(SDE)~\cite{SDE}.
\begin{equation}
    d\boldsymbol{x} = [\boldsymbol{f}(\boldsymbol{x}, \tau) + g(\tau)^{2} \nabla_{\boldsymbol{x}} \log p^{\tau}(\boldsymbol{x}) ] d\tau + g(\tau) d \Bar{\boldsymbol{w}}
\end{equation}
where $\boldsymbol{f}$ is a vector-valued function acting as a drift coefficient, and $g$ is a scalar-valued function acting as the diffusion coefficient of the process, 
$\Bar{\boldsymbol{w}}$ is the reverse-time Wiener process, and 
$\nabla_{\boldsymbol{x}} \log p^{\tau} (\boldsymbol{x})$
is the score function.
In practice, we can use a single time-dependent score model
$\boldsymbol{S}_{\theta}(\boldsymbol{x}, \tau)$
to estimate the score function ~\cite{SDE}:
\begin{equation}
    \mathcal{L}(\theta) = \mathbb{E} [ ||\boldsymbol{S}_{\theta}(\boldsymbol{x}^{\tau}, \tau) - \nabla_{\boldsymbol{x}^{\tau}} \log p^{0\tau}(\boldsymbol{x}^{\tau} | \boldsymbol{x}^{0})||^{2}]
\end{equation}
where the expectation is over diffusion time $\tau$ and 
noised sample 
$\boldsymbol{x}^{\tau} \sim p^{0\tau}(\boldsymbol{x}^{\tau}|\boldsymbol{x}^{0})$,
which is obtained by applying the $\tau$-level perturbation kernel to a clean sample 
$\boldsymbol{x}^{0} \sim p^{data}(\boldsymbol{x}^{0})$.

\section{Method}
To generate trajectory branches, we propose Diffusion-Based Branch Generation(BG), which uses a diffusion model for generation and the Trajectory Value Function (TVF) for guidance.
In the following, we will discuss the detailed pipeline of BG, and introduce the details of model implementation and training.
\subsection{Branch Generation}
We aim to generate trajectory branches that lead to trajectories with higher returns, and expand trajectories of the dataset with the generated branches.
We use the diffusion model to generate the trajectory branches based on the segments sampled from the dataset's trajectories.
We accomplish this by incorporating the trajectory segments into the condition of the diffusion model.
We use the Trajectory Value Function (TVF) to guide the diffusion model to generate branches which can lead to trajectories with higher returns.
The TVF, pre-trained on the dataset, predicts the future return of the trajectory segment used for branch generation.
Since the diffusion model is trained on a limited set of trajectories, guiding returns that significantly exceed the training data may reduce the effectiveness of the guidance.
Thus, the TVF predicts the return based on both the maximum future return and the returns of the trajectories within the dataset.
The generated branches are then used to expand the dataset trajectories. The details of each module are presented below:

\subsubsection{Branch Generating Diffusion Model}
We use a diffusion-based generative model to generate trajectory branches based on the trajectory segments sampled from the dataset.
To achieve consistency between the generated branches and the sampled segment and to condition the branches on the return, we incorporate the sampled segment and its corresponding return into the condition of the diffusion model.
When pre-training the diffusion model, the condition can be represented as:
\begin{equation}
    \begin{aligned}
        c_{t}^{(n)} = 
        \{\langle z_{t-K+1}^{(n)}, z_{t-K+2}^{(n)},...,z_{t}^{(n)} \rangle,R_{t}^{(n)}\} \text{,}
    \end{aligned}
\end{equation}
where $z_{t}^{(n)}=(s_{t}^{(n)},a_{t}^{(n)},r_{t}^{(n)})$. $n \in N$, $N$ denotes the total number of trajectories in the dataset. 
$n$ is the trajectory index of the segment, means the segment is sampled from the $n$-th trajectory in the dataset. $K$ represents the length of the segment, and $R_{t}^{(n)}= \sum_{i=t}^{T}{\gamma^{i-t} r_{t}^{(n)}}$, $\gamma$ is a discount factor. We use $\tau_{t-K+1:t}^{(n)}$ to represent 
$\langle z_{t-K+1}^{(n)}, z_{t-K+2}^{(n)},...,z_{t}^{(n)} \rangle$, which means the trajectory segment.
In the pre-training process, we use the real return of the sampled segment, which is calculated from the corresponding trajectory within the dataset, as the return in the condition.
This makes the generated segment conditioned on the future return.
Based on the condition, the diffusion model generates the succeeding segment:
\begin{equation}
    \langle \tilde{z}_{t+1}, \tilde{z}_{t+2},...,\tilde{z}_{t+H} \rangle=\mathcal{G}_{\theta}(c_{t}^{(n)}) \text{,}
    \label{eq:gene data}
\end{equation}
where $\mathcal{G}_{\theta}$ is the diffusion-based generative model, $H$ denotes the length of the generation horizon. We use $\tilde{\tau}_{t+1:t+H}$ to represent the segment generated by the diffusion model. 
The diffusion model is trained by:
\begin{equation}
    \mathcal{L}_{\theta} = \mathbb{E}_{t,n}||\tau_{t+1:t+H}^{(n)} - \tilde{\tau}_{t+1:t+H}||^{2} \text{,}
    \label{eq:train diff}
\end{equation}
where 
$\tau_{t+1:t+H}^{(n)}$ represents the succeeding segment of $\tau_{t-K+1:t}^{(n)}$ in the real trajectory.

After pre-training, we replace the real return in the condition with the future return predicted by the TVF. This guides the diffusion model to generate trajectory branches that lead to higher returns. The generated branches are then concatenated with the sampled trajectory segments as expansions to the trajectories within the dataset.



\subsubsection{Trajectory Value Function}
To guide the generation of branches that lead to higher return trajectories, we pre-train the Trajectory Value Function (TVF) to predict future returns of the sampled trajectory segments.

Since we only need to estimate the Q-values of state-action pairs within the dataset, inspired by IQL~\cite{IQL}, we adopt a SARSA-style objective to completely avoid the influence of OOD actions during the pre-training, as Eq. \eqref{eq:iql_loss}. 
We predict the Q-value of $(s_{t},a_{t})$, which is the last state action pair of the trajectory segment, as the segment's future return.
However, if the future return predicted by TVF is much higher than returns in the dataset, it will loss its effectiveness as guidance.
To match the return predicted by TVF to the trajectories of the dataset, we constrain it by the maximum corresponding return in the dataset of $(s_{t},a_{t})$:
\begin{equation}
    \min_{\psi} \mathbb{E}_{(s_{t},a_{t}) \in \mathcal{D}}[(V_{\psi}(s_{t})-\max_{n}R^{(n)}(s_{t},a_{t}))^2]\text{,}
\end{equation}
where, $R^{(n)}(s_{t},a_{t})=\sum_{i=t}^{T} {\gamma^{i-t}r(s_{i}^{(n)},a_{i}^{(n)})}$, $n \in N$, means the return of $(s_{t},a_{t})$ in the $n$-th trajectory of the dataset. If the $n$-th trajectory doesn't contain $(s_{t},a_{t})$, we assume $R^{(n)}(s_{t},a_{t})=0$.
Obtaining the of maximum of $R^{(n)}(s,a)$ is difficult when the dataset becomes large.
In practice, we estimate it through expectile regression ~\cite{expectile_regression1,expectile_regression2,IQL},which leads to the following objective:
\begin{equation}
    \mathcal{L}'(\psi)=\mathbb{E}_{(s,a) \sim \mathcal{D},n}[L_{2}^{\tau}(R^{(n)}(s,a)-V_{\psi}(s))] \text{,}
    \label{eq: TVF_loss}
\end{equation}
where $L_{2}^{\tau}(u) = |\tau - \mathds{1}(u<0)|u^{2}$, and $\mathds{1}$ is the indicator function.
During training, we combined it with the original objective of maximum future return from Eq.\eqref{eq:iql_loss}. Thus, TVF is trained by:
\begin{equation}
    \begin{aligned}
        \begin{aligned}
            \mathcal{L}_{V}(\psi)=(1-w)&*\mathbb{E}_{(s,a)\sim \mathcal{D}}[L_{2}^{\tau}(Q_{\phi}(s,a)-V_{\psi}(s))] \\
            &+w*\mathcal{L}'(\psi)
        \end{aligned}\\
        \mathcal{L}_{Q}(\phi)=\mathbb{E}_{(s,a,s') \sim \mathcal{D}} [(r(s,a) + \gamma V_{\psi}(s')-Q_{\phi}(s,a))^{2}]
        \label{eq:train tvf}
    \end{aligned}
\end{equation}
We use $\omega$ to balance the maximum future return and the maximum actual return in the dataset.
When generating branches, we use $Q_{\phi}(s_{t},a_{t})$ as the the future return of the trajectory segment to guide the generation.

\subsection{Model Implementation and Training}
We will discuss the implementation details of the diffusion model, and describe the entire training pipeline in this section. 

The diffusion model is designed to generate trajectory branches, inspired by ~\cite{EDMinspire},we build the diffusion model upon the EDM formulation proposed in ~\cite{EDM}, which is more effective.
The generated branches should be consistent with the environment, which needs high accuracy of the diffusion model. We adopt diffusion transformer~\cite{DiffusionTranformer} which has good performance in vision tasks, to tackle this problem.

The overview of our method is shown in Fig.\ref{fig: method}. 
First, we randomly sample a trajectory segment from the dataset. 
The TVF predicts the future return of the segment. Then we combine the predicted return and the sampled segment together as the condition of the diffusion model, which generates the trajectory branch.
We concatenate the branch with the trajectory segment as an expansion of the trajectory from which the segment was sampled.
To ensure consistency between the generated branches and the preceding trajectory segments, we designed the Branch Filter. 
We filter the generated branches by the continuity of the returns between the branch and the trajectory segment:
\begin{equation}
    \begin{aligned}
        Q_{\phi}^{n}(s_{t},a_{t}) &= \sum_{i=0}^{n-1}{\gamma^{i}{r(s_{t+i},a_{t+i})}} +  \gamma^{n}Q(s_{t+n},a_{t+n}) \text{,} \\
        &\left\lvert  Q_{\phi}(s_{t},a_{t}) - \frac{1}{H} \sum_{i=1}^{H} Q_{\phi}^{i}(s_{t},a_{t}) \right\rvert < \delta \text{.}
    \end{aligned}
    \label{eq: filter}
\end{equation}
$Q_{\phi}^{n}(s_{t},a_{t})$ means the $TD(n)$ target of $(s_{t},a_{t})$ ,$\delta$ denotes the threshold.
Trajectories in the dataset are expanded with branches that satisfy Eq.\eqref{eq: filter}.
Finally, we train DT with the expanded dataset in the original manner.
The entire process is shown in Appendix A.


\begin{table*}[t]
\centering

\begin{tabular}{l|ccc|ccccc}
\toprule
\textbf{Dataset} &\textbf{BC} & \textbf{CQL} & \textbf{IQL} & \textbf{DT} & \textbf{ODT} & \textbf{EDT} & \textbf{QDT} & \textbf{BG+DT}\\
\midrule
maze2d-umaze-v1   & 0.4   & -8.9  & 42.1  & 18.1  &       & 35.8  & 57.3   & \textbf{72.65 $\pm$ 2.0} \\
maze2d-medium-v1  & 0.8   & 86.1  & 34.9  & 31.7  &       & 18.3  & 13.3   & \textbf{143.60 $\pm$ 11.1}\\
maze2d-large-v1   & 2.3   & 23.8  & 61.7  & 35.7  &       & 26.8  & 31.0   & \textbf{83.79 $\pm$ 11.3}\\
\hline
\textbf{Total} & 3.5    & 101.0 & 138.7  & 85.5   &     & 85.9    & 101.6    & \textbf{300.0} \\
\hline
antmaze-umaze-v2          & 68.5  & 94.8  & 84.0 & 57.0  & 53.1  & 67.8  &      & \textbf{71.65 $\pm$ 3.2} \\
antmaze-umaze-diverse-v2  & 64.8  & 53.8  & 79.5 & 51.8  & 50.2  & \textbf{58.3}  &       & 55.50 $\pm$ 4.7 \\
antmaze-medium-play-v2    & 4.5   & 80.5  & 78.5 & 0.8   & 0.0   & 0.0   &     & \textbf{15.62 $\pm$ 4.8} \\
antmaze-medium-diverse-v2 & 4.8   & 71.0  & 83.5 & 0.5   & 0.0   & 0.0   &     & \textbf{13.64 $\pm$ 4.5} \\
\hline
\textbf{Total} & 142.6     & 300.1     & 325.5     & 110.1     & 103.3     & 126.1     &      & \textbf{156.4} \\
\hline
halfcheetah-medium-v2        & 42.6  & 44.0  & 47.4  & 42.6  &\textbf{42.7}  & 42.5  & 42.3  & 42.22 $\pm$ 0.2 \\
halfcheetah-medium-replay-v2 & 36.6  & 45.5  & 44.2  & 36.6  & 40.0  & 37.8  & 35.6  & \textbf{40.35 $\pm$ 0.2} \\
halfcheetah-medium-expert-v2 & 55.2  & 91.6  & 86.7  & 86.8  &       & 91.2  &       & \textbf{92.70 $\pm$ 0.5} \\
hopper-medium-v2             & 52.9  & 58.5  & 66.3  & 67.6  & 67.0  & 63.5  & 66.5  & \textbf{90.58 $\pm$ 2.1} \\
hopper-medium-replay-v2      & 18.1  & 95.0  & 94.7  & 82.7  & 86.6  & \textbf{89.0}  & 52.1  & 85.19 $\pm$ 4.8 \\
hopper-medium-expert-v2      & 52.5  & 105.4 & 91.5  & 107.6 &       & 107.8 &       & \textbf{110.65 $\pm$ 0.4} \\
walker2d-medium-v2           & 75.3  & 72.5  & 78.3  & \textbf{74.0}  & 72.2  & 72.8  & 67.1  & 70.37 $\pm$ 4.3 \\
walker2d-medium-replay-v2    & 26.0  & 77.2  & 73.9  & 66.6  & 68.9  & \textbf{74.8}  & 58.2  & 68.10 $\pm$ 3.1 \\
walker2d-medium-expert -v2   & 107.5 & 108.8 & 109.6 & \textbf{108.1} &       & 107.9 &       & 106.66 $\pm$ 0.9 \\
\hline
\textbf{Total} & 466.7     & 698.5     & 692.6     & 672.6     &      & 687.3     &       & \textbf{706.8} \\
\bottomrule
\end{tabular}
\caption{The average normalized score of different methods. Here $\pm$ denoting the standard deviation. The mean and standard deviation are computed over 5 random seeds. The best results among sequence modeling methods of each setting are marked as \textbf{bold}.}
\label{totalexpresult}
\end{table*}

\section{Results}
We conducted extensive experiments on the Gym, Maze2d, and Antmaze tasks from the D4RL benchmark~\cite{D4RL} to validate the effectiveness of BG. 
In addition, we performed ablation studies on various modules to evaluate the contribution of each component to the overall performance. 
We also visualized the generated trajectory branches to provide a clear demonstration of BG's impact.

\subsection{Experiment Settings}
In this section, we will introduce the experimental settings and baseline algorithms.

\paragraph{Evaluation Environments}
We evaluate BG across various domains within the D4RL benchmark, including Maze2d, Antmaze, and MuJoCo tasks. 
The Maze2d tasks feature three map types: umaze, medium, and large, with progressively increasing size and complexity.
The Antmaze tasks include two map types: umaze and medium, each with two task variations: play and diverse.
On MuJoCo tasks, we evaluate BG on three environments: Halfcheetah, Hopper, and Walker2d. Each environment contains three types of datasets: Medium, Medium-replay, Medium-expert, where the Medium-expert dataset consists of both expert and sub-optimal data, and the Medium and Medium-replay datasets are collected by an unconverged SAC policy~\cite{SAC} interacts with the environment.
\paragraph{Baselines}
We compare BG+DT with representative methods from offline RL, imitation learning, and sequence modeling.  The Offline RL methods include Conservative Q-Learning (CQL)~\cite{CQL}, and Implicit Q-Learning (IQL)~\cite{IQL}. For imitation learning, we compare BG+DT with Behavior Cloning (BC) ~\cite{BC}. Sequence modeling methods include Online Decision Transformer (ODT)~\cite{intro_ODT}, Elastic Decision Transformer (EDT)~\cite{intro_EDT}, and Q-learning Decision Transformer (QDT)~\cite{QDT}. 
For a fair comparison, we evaluate ODT in its offline version.
We also compare with the original DT without BG processing. Please refer to Appendix C for detail.

\paragraph{Implementation Details} 
For the diffusion model, the generation horizon is set to 10 for most of the tasks in the Gym, Maze2d, Antmaze tasks. 
The length of the condition segment is the same as the generation horizon. 
The denoising step is set to 10.
We set the horizon of DT to 20, the same length as the trajectory branches, to take advantage of these branches effectively.
Please refer to Appendix B for detail.


\subsection{Results on D4RL Benchmark}
We conduct a comprehensive evaluation of BG on the D4RL benchmark, covering both dense reward tasks, including Gym, Maze2d and sparse reward task AntMaze.
Without processing with BG, DT performs relatively well on the Gym dataset, with performance comparable to offline RL methods.
However, DT struggles on Maze2d and Antmaze tasks, where it is more constrained by sub-optimal trajectories.
After processing with BG, DT's performance on the Maze2d and Antmaze tasks shows significant improvement.

\paragraph{Results on Dense Reward Dataset}
The results of BG+DT on Gym and Maze2d tasks are presented in Table.\ref{totalexpresult}.
Previous sequence modeling methods have focused primarily on Gym tasks, where their performance is comparable to state-of-the-art offline algorithms.
However, on Maze2d tasks that is more challenging, DT and other sequence modeling methods perform poorly. 
But after expanding datasets with generated branches on Maze2d task, DT's performance improves significantly.

\begin{table}[t]
\centering
\begin{tabular}{l|ccc}
\toprule
\textbf{Dataset} & \textbf{DT} &\textbf{with-TVF} & \textbf{no-TVF} \\
\midrule
maze2d-umaze-v1  & 18.1 & 72.65 & 51.11 $\pm$ 15.0 \\
maze2d-medium-v1 & 31.7 & 143.60 & 131.52 $\pm$ 27.7 \\
maze2d-large-v1  & 35.7 & 83.79 & 66.07 $\pm$ 7.3 \\
\bottomrule
\end{tabular}
\caption{The impact of TVF. No-TVF means guiding the diffusion model with the value function that does not consider the real return from the dataset which degrades to IQL Q-function.}
\label{TVF-ablation}
\end{table}

\begin{table}[t]
\centering
\begin{tabular}{l|ccc}
\toprule
\textbf{Dataset} & \textbf{DT} &\textbf{with-Filter} & \textbf{no-Filter} \\
\midrule
maze2d-umaze-v1  & 18.1 & 72.65 & 59.16 $\pm$ 10.2 \\
maze2d-medium-v1 & 31.7 & 143.60 & 111.32 $\pm$ 26.5 \\
maze2d-large-v1  & 35.7 & 83.79 & 57.69 $\pm$ 6.3 \\
\bottomrule
\end{tabular}
\caption{The impact of Branch Filter.}
\label{ablation_filter}
\end{table}

\paragraph{Results on Sparse Reward Dataset}
The Antmaze task features sparse rewards, with $r=1$ when reaching the goal. Both the medium-diverse and medium-play datasets do not contain complete trajectories from the starting point to the goal.
On these datasets, sequence modeling methods struggle for lack of the ability to learn policies that can move across trajectories.
As shown in Table.\ref{totalexpresult}, sequence modeling methods are basically unable to reach the goal on the medium-diverse and medium-play datasets.
But after adding branches, the sub-optimal trajectories can be connected by branches, which helps DT learn policies that combine different sub-optimal trajectories to reach the goal.
BG+DT can learn policies that have the ability to reach the goal, which is a significant improvement over sequence modeling methods.
\subsection{Ablation Study}
To investigate the impact of the modules in BG, we apply ablation studies on the TVF and Branch Filter modules. 
Since our method shows the greatest improvement on the Maze2d task, and it most effectively reflects the impact of our method, the ablation studies are primarily conducted on the Maze2d task.
\paragraph{Trajectory Value Function Ablation}
In the TVF, we incorporate an actual return loss, as seen in Eq.\eqref{eq: TVF_loss}, to consider the consistency between the predicted return and the actual return in the dataset.
We compare the performance of TVF with and without the actual return loss as the guidance for the diffusion model.
As Table.\ref{TVF-ablation} illustrated, removing the actual return loss results in a noticeable decline in the method's performance.

\begin{figure}[t]
    \centering
    \includegraphics[width=0.8\linewidth]{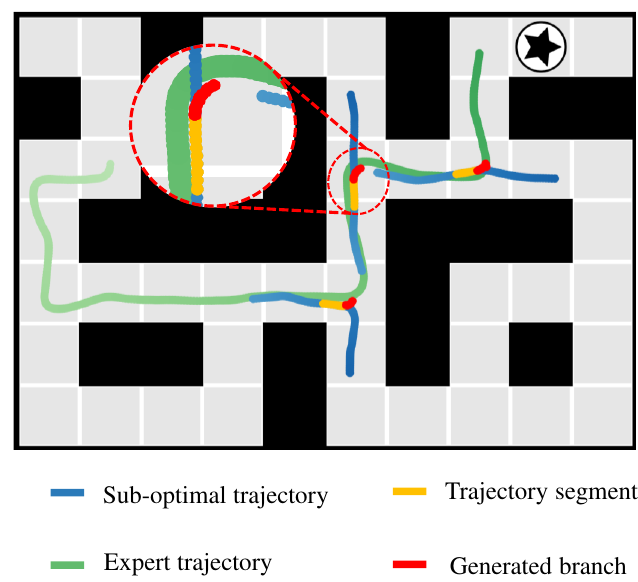}
    \caption{Branch demonstration on Maze2d-large map. The black pentagram represents the goal. The direction in which the color changes from light to dark is the direction of the trajectory.}
    \label{fig:maze2d_visual}
\end{figure}

\begin{figure}[t]
    \centering
    \begin{subfigure}[b]{\linewidth}
        \centering
        \includegraphics[width=1.0\linewidth]{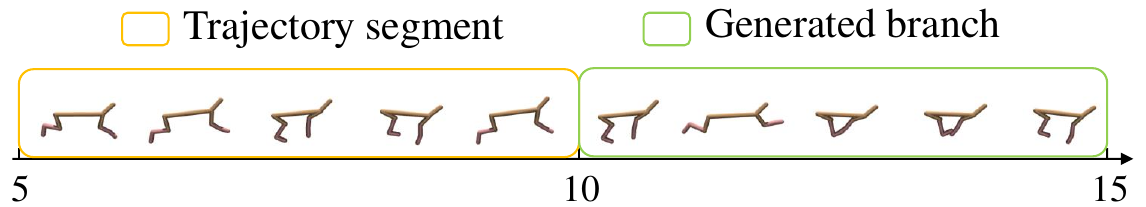}
        \caption{Halfcheetah}
        \label{fig:sub1}
    \end{subfigure}
    \hfill
    \begin{subfigure}[b]{\linewidth}
        \centering
        \includegraphics[width=1.0\linewidth]{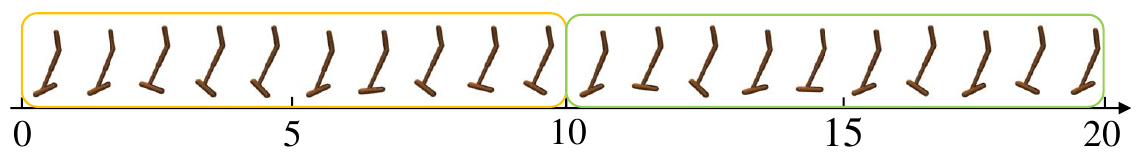}
        \caption{Hopper}
        \label{fig:sub2}
    \end{subfigure}
    \hfill
    \begin{subfigure}[b]{\linewidth}
        \centering
        \includegraphics[width=1.0\linewidth]{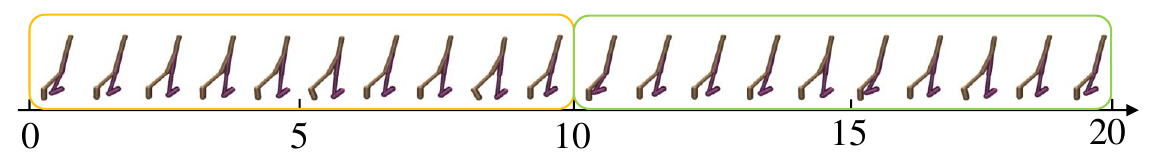}
        \caption{Walker2d}
        \label{fig:sub3}
    \end{subfigure}
    
    \caption{Branches demonstration on Gym tasks. The length of the branches in Halfcheetah is 20, but we only show the branch from 5 to 15 to narrow the width.}
    \label{fig:DMC_demonstration}
\end{figure}

\paragraph{Branch Filter Ablation}
Though the branches can be generated with the techniques mentioned previously, their quality varies. 
To ensure continuity between the generated segments and the preceding ones, we use the consistency of their returns to filter the generated segments.
We compared the performance with and without the Branch Filter module of our method, and the results are presented in Table.\ref{ablation_filter}. 
We can observe that without the Branch Filter module, the performance of BG significantly declined.

\subsection{Visualize Demonstration}

We visualized the generated trajectory branches and the trajectory segments on Maze2d and Gym tasks to investigate whether the generated trajectory branches lead to better trajectories and are consistent with the dynamics of the environment.
As shown in Fig.\ref{fig:maze2d_visual}, we simultaneously displayed the expert trajectories, the sub-optimal trajectories, and the trajectory branches with their base segment in the same image, demonstrating that the trajectory branches based on sub-optimal trajectories can branch off and lead to the good trajectories. 
The generated branch can apparently branch off from the sub-optimal trajectory, and lead to the expert trajectory.
For the agent in Maze2d task has velocity, if it wants to branch off from the sub-optimal trajectory, it should first slow down and turn the direction gradually, resulting in the branch may not be so obvious in visual.

To investigate whether the trajectory branches are consistent with the dynamics of the environment, we visualize the robot states at each time step of the branches of HalfCheetah, Walker2d, and Hopper.
We can observe that the branches generated by the diffusion model are quite natural and roughly match the locomotion patterns of these robots in Fig.\ref{fig:DMC_demonstration}.


\section{Related Works}
\paragraph{Offline RL} Offline Offline RL~\cite{intro_offlinebegin1,relwork_offlinebegin1} aims to learn a policy exclusively from a previously collected static dataset. Since the learned policy may diverge from the behavior policy that generated the dataset, offline RL algorithms primarily focus on mitigating the impact of distributional shift~\cite{intro_offlinebegin1,relwork_offlinebegin1}.
One approach to addressing this issue is through policy constraints, which enforce the learned policy to remain close to the behavior policy. This proximity is typically ensured using batch constraints~\cite{intro_offlinebegin1}, KL divergence~\cite{policyconstrain_kl}, Maximum Mean Discrepancy (MMD)~\cite{policyconstrain_mmd}, or Mean Squared Error (MSE) constraints~\cite{policyconstrain_mse}.
Another class of methods tackles the out-of-distribution (OOD) problem through value function regression. For instance, Conservative Q-Learning (CQL)~\cite{CQL} penalizes the Q-values of OOD actions, while Implicit Q-Learning (IQL)~\cite{IQL} addresses the OOD issue by avoiding explicit estimation of OOD actions through implicit learning techniques.
Additionally, one-step RL methods~\cite{OnesetpRL1} perform in-sample Bellman updates to accurately estimate the Q-function, followed by a single policy improvement step to derive the optimal policy.
Imitation learning approaches~\cite{imitationlearning1,imitationlearning2,imitationlearning3} also contribute to offline RL by learning policies that imitate optimal behaviors while filtering out sub-optimal actions.

\paragraph{Sequence Modeling in Offline RL} 
DT~\cite{intro_DT} incorporates return as part of the sequence to predict the optimal action, breaking away from the classic RL paradigm and directly addressing OOD problems.
Some works have made improvements to DT.
Trajectory Transformer (TT)~\cite{TT} models distributions over trajectories using a transformer architecture and incorporates beam search as a planning algorithm.
QDT~\cite{QDT} relabels the ground-truth return-to-go with estimated values to enhance trajectory recombination.
EDT~\cite{intro_EDT} adjusts the history length maintained in DT to facilitate trajectory stitching.
ODT~\cite{intro_ODT} blends offline pretraining with online fine-tuning within a unified framework.

\paragraph{Data Augmentation in Offline RL}
Previous research has also explored methods to augment offline RL datasets.
A class of methods enhances datasets through trajectory rollout~\cite{rollout1, rollout2_TATU, rollout3_CABI}.
These methods require the construction of a dynamic model and an associated rollout policy to generate the trajectories. 
TATU~\cite{rollout2_TATU} uses a forward dynamic model for trajectory rollout, while ensuring the uncertainty of the rolled-out trajectories through truncation mechanisms. 
CABI~\cite{rollout3_CABI} uses bidirectional dynamic models for trajectory rollout and performs a double check on the rolled-out trajectories to improve their accuracy.
However, these dynamic model-based rollout methods are limited to short-distance rollouts. When the rollout distance becomes too long, the quality of the trajectories is difficult to guarantee, making them unsuitable for the sequence modeling approach used by DT.
Some methods enhance datasets by generating new transition pairs~\cite{tuple_simulation1}.
SER~\cite{tuple_simulation1} uses a diffusion model to learn from the dataset and then generate new transition pairs. 
However, this method cannot generate sequences and thus does not support sequence modeling. 
DiffStitch~\cite{Diffstitch} performs data augmentation by randomly concatenating two trajectories from the dataset. But, since the trajectories are randomly selected from the dataset, it is difficult to ensure the quality of the concatenated segments. 
TS~\cite{TS} concatenates trajectories by searching for the next better state. Although this method can also connect to better trajectories, it lacks generation capability and is overly dependent on the value function and the dynamic model.

\section{Discussion}
There are some previous works ~\cite{rollout2_TATU,rollout3_CABI,tuple_simulation1,TS,Diffstitch} that also augment the dataset. Our method is different from them in the following aspects. First, BG expands the dataset with trajectory segments, which is especially for sequence modeling methods. 
Most of the previous works ~\cite{rollout2_TATU,rollout3_CABI,tuple_simulation1} can not generate long segments with high accuracy, which can not be used on sequence modeling methods.
Second, BG bridges the sub-optimal trajectories and better trajectories by generation. Previous works ~\cite{TS} accomplish this by stitching trajectories in the dataset, which lack the bridging ability.
Third, in BG, which trajectory the branch is connected to is decided implicitly by the maximum future return and the dataset distribution through guiding the diffusion model with a value function. Compared with previous works~\cite{Diffstitch}, which connect trajectories randomly, BG is more aligned with the dataset distribution.

As observed in Table.\ref{totalexpresult}, BG improves DT mainly in Maze2d and Antmaze tasks, the performance improvement in Gym tasks is less significant. 
We think this is mainly because, in maze tasks, the trajectories that reach the goal and those that do not are separated by a large spatial distance, making the transition from sub-optimal trajectories to expert demonstrations more difficult than in Gym tasks. 
In Gym tasks, the agent can transition from a sub-optimal trajectory to an expert trajectory within a few steps. 
But in Maze2d and Antmaze tasks, the transition may require a long sequence of actions, making it difficult for DT to generate without corresponding training data.
Thus, trajectory branches are more effective in Maze2d and Antmaze tasks, as they provide the agent with more 
opportunities to move from sub-optimal trajectories to better ones.

In Fig.\ref{fig:intro}, the trajectory branch has a segment that coincides with the high-return trajectory. We do this because even though the trajectory branch is fully generated, the diffusion model tends to generate segments that overlap with the high-return trajectory when the generated state is near the high-return trajectory. If the generated branch reaches a better trajectory before it ends, the subsequent part of the branch tends to overlap with the better trajectory. In this situation, the generation of the subsequent part is guided by the same return as the better trajectory and the condition segment partially overlaps with it, causing the diffusion to generate the branch's subsequent part along the better trajectory.

Although BG performs well in the experiments, it has some limitations. The branch is generated by the diffusion model, although we use TVF as guidance, we cannot fully control the generation quality. We have removed some bad branches by some filtering techniques, but there may still be some bad branches remaining in the dataset.
The generation horizon of the diffusion model is limited. Although BG's generation accuracy of long trajectory segment is much higher than trajectory rollout methods, we can not generate the entire trajectory with high accuracy. The diffusion model is trained on a limited dataset, its generation quality will deteriorate if the generation horizon becomes too long. 

\section{Conclusion and Future Work}
In this paper we introduce BG, which enhances DT by expanding the trajectories within the dataset with trajectory branches. Our method prevents DT from converging to sub-optimal trajectories and provides DT with more opportunities to learn policies that can move to better trajectories.
Empirical evaluations conducted on the D4RL benchmarks demonstrate a significant performance boost for DT.

For future work, applying BG to image-based tasks with large datasets will be promising.  Given the abundance of available image data, pre-training the Diffusion model on these datasets could lead to high accuracy, as diffusion models have demonstrated strong performance in visual tasks.

\bibliography{arxiv_2.2}

\end{document}